\documentclass[journal,hidelinks]{IEEEtran}
\IEEEoverridecommandlockouts

\usepackage{times}
\usepackage[utf8]{inputenc} %
\usepackage[T1]{fontenc}    %
\usepackage{url}            %
\usepackage{booktabs}       %
\usepackage{nicefrac}       %
\usepackage{microtype}      %
\usepackage{graphics,color}

\usepackage{algorithm}
\usepackage{enumitem}

\usepackage{amsfonts}       %
\usepackage{amsmath}       %
\usepackage{amssymb}

\newcommand{\Figref}[1]{Figure~\ref{#1}}  %
\newcommand{\figref}[1]{Fig.~\ref{#1}}    %
\newcommand{\algoref}[1]{Alg.~\ref{#1}}    %

\newcommand{\tabref}[1]{Tab.~\ref{#1}}

\newcommand{\secref}[1]{Sec.~\ref{#1}} %
\newcommand{\suppref}[1]{Suppl.~\ref{#1}}

\usepackage{xspace}
\makeatletter
\DeclareRobustCommand\onedot{\futurelet\@let@token\@onedot}
\def\@onedot{\ifx\@let@token.\else.\null\fi\xspace}
\makeatother

\newcommand{\eg}{e.g\onedot}

\usepackage{xr-hyper}
\makeatletter
\newcommand*{\addFileDependency}[1]{%
  \typeout{(#1)}
  \@addtofilelist{#1}
  \IfFileExists{#1}{}{\typeout{No file #1.}}
}
\makeatother

\usepackage{xcolor}
\definecolor{ourblue}{rgb}{0.368,0.507,0.71}
\definecolor{ourorange}{rgb}{0.881,0.611,0.142}
\definecolor{ourgreen}{rgb}{0.56,0.692,0.195}
\definecolor{ourred}{rgb}{0.923,0.386,0.209}
\definecolor{ourviolet}{rgb}{0.528,0.471,0.701}
\definecolor{ourbrown}{rgb}{0.772,0.432,0.102}
\definecolor{ourlightblue}{rgb}{0.364,0.619,0.782}
\definecolor{ourdarkgreen}{rgb}{0.572,0.586,0.}

\definecolor{ourred2}{rgb}{0.84,0.15,0.16}
\definecolor{ourorange2}{rgb}{1,0.5,0.05}
\definecolor{ourblue2}{rgb}{0.12,0.47,0.71}
\definecolor{ourgreen2}{rgb}{0.17,0.63,0.17}
\definecolor{ourgray2}{rgb}{0.82,0.82,0.82}
\definecolor{ourgreen2}{rgb}{0.29,0.62,0.224}

\usepackage{subcaption}
\graphicspath{../figs/}
\newcommand{\myparagraph}[1]{%
    \par\noindent\hspace*{1em}\textbf{#1}\hspace{0.1em}%
}

\newcommand{\limittorque}{\ensuremath{\tau_{i}^{\mathrm{lim}}}\xspace}
\newcommand{\smooth}{\ensuremath{(u - u_{\mathrm{prev}})^{2}}\xspace}
\newcommand{\weight}{\ensuremath{w}}
\newcommand{\cost}{\ensuremath{c}}

\newcommand{\fgrf}{\ensuremath{F_{j}^{\mathrm{GRF}}}\xspace}
\newcommand{\numbermus}{\ensuremath{N_{\mathrm{active}}}\xspace}

\newcommand{\hyfydy}{Hyfydy\xspace}
\newcommand{\mujoco}{MuJoCo\xspace}
\newcommand{\planarmodel}{\textit{H0918}\xspace}
\newcommand{\threedmodel}{\textit{H1622}\xspace}
\newcommand{\complexmodel}{\textit{H2190}\xspace}
\newcommand{\myoleg}{\textit{MyoLeg}\xspace}

\usepackage{pifont}%
\newcommand{\cmark}{\ding{51}}%
\newcommand{\xmark}{\ding{55}}%

\usepackage{cite}
\usepackage{amsmath,amssymb,amsfonts} 
\usepackage{algpseudocode}
\usepackage{graphicx}
\usepackage{textcomp}
\usepackage{xcolor}
\usepackage{multicol}

\usepackage{hyperref}

\usepackage{capt-of,etoolbox}

\makeatletter
\apptocmd\@maketitle{{\myfigure{}\vspace{-20pt}}}{}{}
\makeatother

\begin{document}
\newcommand\myfigure{%
\centering     
\setcounter{figure}{0} %
        \includegraphics[height=0.2\textheight]{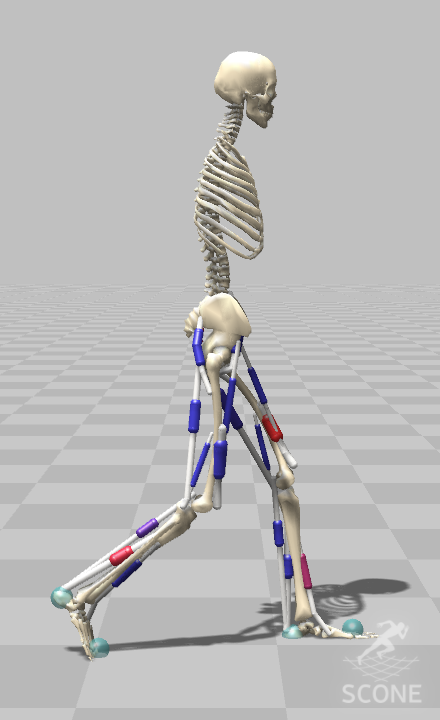}
        \includegraphics[height=0.2\textheight]{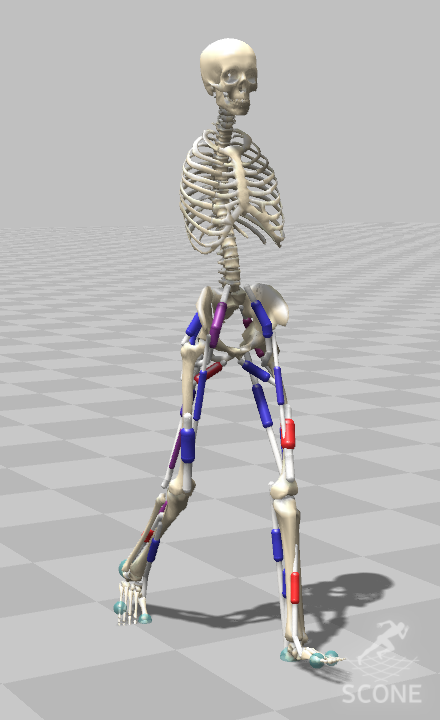}
        \includegraphics[height=0.2\textheight]{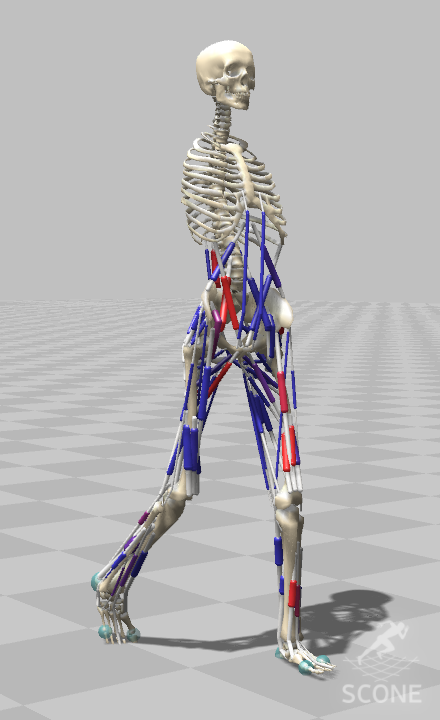}
        \includegraphics[height=0.2\textheight]{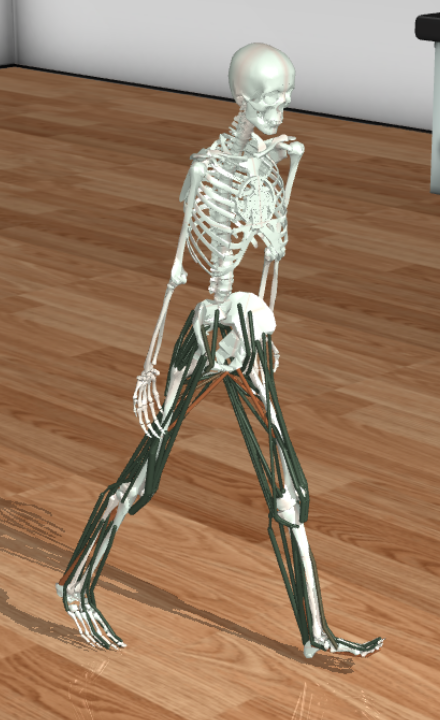}
        \includegraphics[height=0.2\textheight]{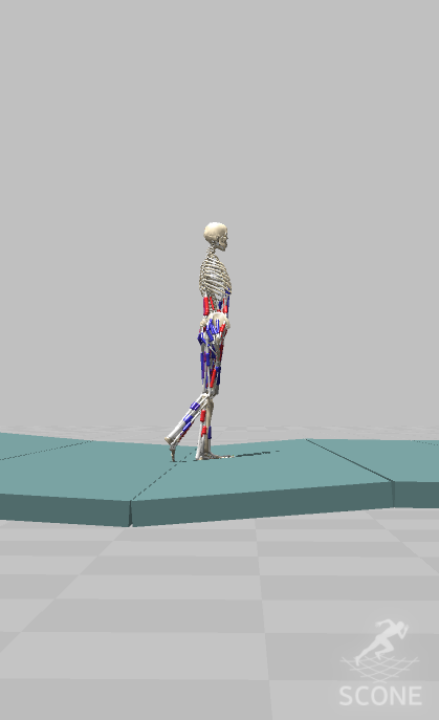}\\
\captionof{figure}{{\bf{We achieve robust and energy-efficient natural walking with RL on a series of human models}} Left to right: \planarmodel, \threedmodel, \complexmodel, \myoleg and an uneven terrain environment. Videos: \href{https://sites.google.com/view/naturalwalkingrl}{\color{blue}https://sites.google.com/view/naturalwalkingrl}}
\label{fig:models}
}
\title{Natural and Robust Walking using Reinforcement Learning without Demonstrations in High-Dimensional Musculoskeletal Models}
\author{Pierre Schumacher$^{1,2}$, Thomas Geijtenbeek$^{3}$, Vittorio Caggiano$^{4}$, Vikash Kumar$^{4}$, Syn Schmitt$^{5}$,\\ Georg Martius$^{1,6}$, Daniel F. B. Haeufle$^{2,7}$
\thanks{$^{1}$Max-Planck Institute for Intelligent Systems, Tübingen, Germany}
\thanks{$^{2}$Hertie Institute for Clinical Brain Research and Center for Integrative Neuroscience, Tübingen, Germany}
\thanks{$^{3}$Goatstream, The Netherlands}
\thanks{$^{4}$Meta AI, New York, USA}
\thanks{$^{5}$Institute for Modelling and Simulation of Biomechanical Systems, University of Stuttgart, Germany}
\thanks{$^{6}$Department of Computer Science, University of Tübingen, Germany}
\thanks{$^{7}$Institute of Computer Engineering, Heidelberg University, Germany}
}
\markboth{Preprint}{Schumacher \MakeLowercase{\textit{(et al.)}:Title}}
\maketitle

\begin{abstract}

Humans excel at robust bipedal walking in complex natural environments. In each step, they adequately tune the interaction of biomechanical muscle dynamics and neuronal signals to be robust against uncertainties in ground conditions.
However, it is still not fully understood how the nervous system resolves the musculoskeletal redundancy to solve the multi-objective control problem considering stability, robustness, and energy efficiency.
In computer simulations, energy minimization has been shown to be a successful optimization target, reproducing natural walking with trajectory optimization or reflex-based control methods. However, these methods focus on particular motions at a time and the resulting controllers are limited when compensating for perturbations. %
In robotics, reinforcement learning~(RL) methods recently achieved highly stable (and efficient) locomotion on quadruped systems, but the generation of human-like walking with bipedal biomechanical models has required extensive use of expert data sets. This strong reliance on demonstrations often results in brittle policies and limits the application to new behaviors, especially considering the potential variety of movements for high-dimensional musculoskeletal models in 3D. Achieving natural locomotion with RL without sacrificing its incredible robustness might pave the way for a novel approach to studying human walking in complex natural environments.
\end{abstract}
\section{Introduction}
The aim of this study is to demonstrate that RL methods can generate robust controllers for musculoskeletal models. The novelty of our work is that we do not try to achieve human-like behavior with RL by relying on kinematic data, but only on biologically plausible objectives in combination with realistic biomechanical constraints embedded into simulation engines. These evolutionary priors have the potential to be general enough to allow for the reproduction of natural gait, similar to the achievements of reflex control policies, but with the potential for generating diverse and robust behaviors under many different conditions.%

We specifically propose a reward function restricted to metrics that are considered plausible objectives for biological organisms, while using experimental human data only to modify the relative importance of the different metrics, similar to \cite{inversecostfunctiontuning2011, amp2021, costfunctionvaries2023}. This goes beyond previous works applying RL to biomechanical models which either study low-dimensional systems~\cite{wengnatwalking2021}, make use of expert data\cite{gaitnet2022} during training or learn unrealistic movements~\cite{accelerated2022, schumacher2023deprl}. We propose a reward function based only on walking speed, joint pain, and muscle effort, achieving periodic gaits that resemble human walking kinematics and ground reaction forces~(GRF) closer than comparable RL approaches \cite{Wang2012, learn2run2018, songDeepReinforcementLearning2020a, SARBerg2023}. Furthermore, the learning approach generated walking in 4 different models and 2 simulation engines of differing biomechanical complexity and accuracy with an identical training protocol and without changing the reward function.

The simpler 2D and 3D models are comparable in complexity and almost reach the naturalism of existing optimal control- and reflex-based frameworks \cite{geyer2010,Song2013,Geijtenbeek2013}. While their 80 or 90 muscle counterparts are substantially more challenging to control, our approach still achieved gaits with kinematics and GRFs similar to experimental human data, albeit with more artifacts, potentially related to biomechanical modeling accuracy. Achieving gaits in these complex models is a step towards applications in rehabilitation, neuroscience, and computer graphics requiring high-dimensional models in complex environments. Striking is the robustness of the learned controllers exhibiting diverse stabilization strategies when faced with dynamic perturbations to an extent unseen in previous reflex-based controllers \cite{geyer2010,Song2017a, Haeufle2018a, Ramadan2022a, Schreff2022a}. As the used reward terms are considered plausible objectives for biological organisms, the general approach may also be applicable to different movements. Therefore, we believe that this approach is a useful starting point for the community showing that RL \textbf{is} a viable candidate to investigate the highly robust nature of complex human movements.\looseness-1

\section{Results}

Our framework is built upon the recently published DEP-RL~\cite{schumacher2023deprl} approach to learning feedback controllers for musculoskeletal systems. DEP-RL has been shown to achieve robust locomotion in several tasks, including running with a high-dimensional~(120 muscles) bipedal ostrich model, by proposing a novel exploration scheme for overactuated systems. The learned behaviors, however, still exhibited unnatural artifacts, such as large co-contraction levels and excessive bending of several joints. 

Here, we extend that work by introducing an adaptive reward function that accounts for biologically plausible incentives. These incentives result in gaits that resembling human walking much closer. Furthermore, the reward function is general enough to generate gaits across several models with up to 90 muscles in two and three dimensions and in simulators of differing biomechanical modeling accuracy \textbf{without} changes in the weighting of the incentives in the reward function. Only the network size was decreased for the low-dimensional models to benefit from the computational speed up.

\subsection{Reward function}
Building on previous work on gait optimization~\cite{Geijtenbeek2013}, we found that a natural gait can be achieved with RL by using objectives that incentivize:
\begin{enumerate}
    \item learning to maintain a given speed without falling down,
    \item minimizing effort, and
    \item minimizing pain.
\end{enumerate}
Thus, our reward function contains three main terms:
\begin{equation}
    r = r_{\mathrm{vel}} - \cost_{\mathrm{effort}} - \cost_{\mathrm{pain}}.
\end{equation}
The first term specifies the external task the agent should solve. As we want the agent to move at a walking pace while keeping its balance, we chose the following objective:
\begin{equation}
r_{\mathrm{vel}} =\begin{cases}
      \exp{\left[- (v - v_{\mathrm{target}})^{2}\right]}\quad\mathrm{if}\, v < v_{\mathrm{target}}\\
     1\quad\mathrm{otherwise},
     \end{cases}
\end{equation}
where $v$ is the center-of-mass velocity and the target velocity $v_{\mathrm{target}}$ is chosen to be $1.2$\,\nicefrac{m}{s}, which is close to the average energetically optimal human walking speed~\cite{efficientspeed2007}.
The velocity reward is constant above the target velocity to improve the optimization of the auxiliary cost terms, inspired by a recent study on reward shaping in robotics%
~\cite{rudin2022skills}. 

Important for achieving natural human walking is the use of minimal muscle effort, as the literature suggests that energy efficiency is a key component of human locomotion\cite{dajiroenergyefficient2015, raffaltenergyefficient2017}: 
\begin{equation}
   \cost_{\mathrm{effort}} = \alpha({t})\, a^{3} + \weight_{1} \, \smooth + \weight_{2}\,\numbermus
\end{equation}
where the first term penalizes muscle activity $a$\cite{ackermann2010},  the second term incentivizes smoothness of muscle excitations $u$, and the third term \numbermus incentivizes a small number of active muscles (penalizing activity exceeding a certain value). 

From a technical standpoint, it proved challenging to effectively minimize muscle activity. Using a strong cost scale that leads to energy-efficient walking later in training, causes a performance collapse when enabled from the start. We, therefore, chose an approach rooted in constrained optimization\cite{zahavy2023discovering}. We propose an adaptation mechanism for the weighting parameter $\alpha({t})$, increasing the weight only when the agent performs well in the main task ($r_{\mathrm{vel}}$) and decreasing it when this constraint is violated. Concretely, we measure the performance by the task return. The details are provided in \algoref{alg:effort}, we marked the constrained optimization in blue.

This adaptive learning mechanism can be applied to each model and removes the need for hand-tuning of schedules. A change in reward function over time could, however, destabilize learning, as previously collected environment transitions are not reflective of the current effort cost anymore\cite{2021pebble}. We, therefore, monitor the performance of the policy in the current environment, while the effort cost is only applied the moment when data is sampled from the replay buffer. This relabeling of previously collected data ensures that our off-policy algorithm can make efficient use of the full replay buffer.

\begin{algorithm}
\caption{Effort weight adaptation.}
\label{alg:effort}
\begin{algorithmic}
\Require{threshold $\theta$, smoothing $\beta$, change in adaptation rate $\Delta\alpha$, decay term $\lambda\in[0,1]$}
\State $r_{\mathrm{mean}} \gets 0$,  $\alpha_{t} \gets 0$, $\mathrm{s}_{\mathrm{mean}} \gets 0$

\While{True}
\State $r \gets \mathrm{train\_episode}()$ \Comment{return from episode}
\State $r_{\mathrm{mean}} \gets \beta\,r_{\mathrm{mean}} + (1-\beta)\,r $
\If{$r_{\mathrm{mean}} > \theta$ \textbf{and} $\mathrm{s}_{\mathrm{mean}} < 0.5$}
\State $\Delta\alpha \gets \lambda \cdot \Delta\alpha$ \Comment{performance newly high}
\State  \Comment{slow down adaptation}
 \ElsIf{$r_{\mathrm{mean}} > \theta$ \textbf{and} $\mathrm{s}_{\mathrm{mean}} > 0.5$}
   \State {\color{blue}$\alpha_{t+1} \gets \alpha_{t} + \Delta\alpha$} \Comment{performance high for long}
\Else
\State {\color{blue}$\alpha_{t+1} \gets \alpha_{t} - \Delta\alpha$} \Comment{performance too low}
\EndIf
\State $c_{\mathrm{target}} \gets \begin{cases}1 & \textbf{if } r_{\mathrm{mean}} > \theta \\ 
0 & \textbf{otherwise}
\end{cases}$

\State $c_{\mathrm{mean}} \gets \beta c_{\mathrm{mean}} + (1-\beta)\, c_{\mathrm{target}}$
\EndWhile
\end{algorithmic}

\end{algorithm}

The third term $\cost_{\mathrm{pain}}$ is necessary to prevent unnatural optima. One striking example is the over-use of mechanical forces of the joint limits (\eg massive knee over-extension) to keep a straight leg while minimizing muscle activity. As this is clearly unnatural behavior, we include objectives that account for the notion of pain:
\begin{equation}
    \cost_{\mathrm{pain}} = \weight_{3}\, \sum_{i} \limittorque + \weight_{4}\, \sum_{j}\fgrf, 
\end{equation}
where $\limittorque$ is the torque with which the joint angle limit of joint $i$ is violated (joint-limit pain) and \fgrf is the vertical ground reaction force~(GRF) for foot $j$ (joint-loading pain). We only penalize GRFs if they exceed $1.2$ times the model's body weight~\cite{grfwalking1989, Geijtenbeek2019}, such that all pain cost terms vanish close to the natural gait and do not further bias the solution. 

We tuned the cost term weights $\omega_{i}$ for $i\in\{1,...,4\}$ by first separating the kinematic data into gait cycles for each leg, starting and ending when the respective foot touches the ground. The resulting data is then averaged over all gait cycles recorded from both legs. The average trajectory is finally compared to its equivalent obtained from experimental human data. The experimental match, defined as the fraction of the gait cycle for which the average simulated trajectory overlaps within the standard deviation of experimental data, serves as an optimization metric for our cost terms. We note that the coefficients are identical across all joints and muscles, and stress that no human data was used \textbf{during} the learning process, but only to find weighting coefficients. This procedure is similar to \cite{inversecostfunctiontuning2011}, with the difference that we search for values that work across a range of models, instead of optimally for one model.

Finally, we initialize the models with a randomized initial state that starts with one elevated leg, while we also clip all muscle excitations to lie between $0$ and $0.5$ to further reduce muscle effort and mitigate asymmetries caused by the initial state distribution.

\subsection{Models}
With the reward function and the RL approach described above, we are able to learn robust control policies for several models of human walking, with varying complexity, and across two different simulation engines with different levels of biomechanical accuracy  (see \figref{fig:models}):
\begin{table}
\caption{All used models. Trunk means that the trunk and the pelvis of the model can move separately. Toes means that the toes and the rest of the foot can move separately. The designation 3D marks models that can walk in full 3D, as opposed to planar movements.}
\centering
\begin{tabular}{@{}c@{\ }ccccccc@{}}
 Model & \# DOFs  & \# muscles & 3D & trunk & toes &engine\\
 \toprule
 H0918 &  9  & 18 & \xmark & \xmark & \xmark &\hyfydy \\
 
 H1622  & 16  & 22  &\cmark&\xmark & \xmark &\hyfydy \\
 H2190  & 21  & 90  & \cmark & \cmark & \xmark &\hyfydy \\
 MyoLeg  & 20 & 80 & \cmark & \xmark & \cmark & \mujoco\\
\bottomrule
\end{tabular}
\label{tab:models}
\end{table}
\myparagraph{H0918} A planar \hyfydy model with 9 degrees-of-freedom (DOFs) and 18 muscles, based on \cite{Delp1990}.
\myparagraph{H1622} A 3D \hyfydy model with 16 DOFs and 22 muscles, based on \cite{Delp1990}.
\myparagraph{H2190} A 3D \hyfydy model with 21 DOFs and 90 muscles, and articulation between the otherwise rigid pelvis and torso, based on \cite{Delp1990, Rajagopal2016, Christophy2012}.
\myparagraph{MyoLeg} A 3D \mujoco model with 20 DOFs and 80 muscles, based on \cite{Rajagopal2016}. As for the \planarmodel and \threedmodel models, the pelvis and torso are one rigid body part, while each foot contains articulated toes (all five toes are joined into one body segment). See \tabref{tab:models} for a summary of the models.

\subsection{Simulation engine}
The simulation engines used for each model are indicated in the description and are either: \textit{a)} \hyfydy~\cite{Geijtenbeek2021}, which was used via the SCONE Python API~\cite{Geijtenbeek2019}, or \textit{b)} \mujoco, which was used via the MyoSuite~\cite{MyoSuite2022} environment. We chose these two engines, to highlight the versatility of our approach but also to bridge two communities: biomechanics and RL. 

\hyfydy is an engine built for biomechanical accuracy. It is closely related to the well-established OpenSim~\cite{Seth2018} framework, matching its level of detail in muscle and deformation-based contact-force models while providing increased computational performance. \mujoco is a fast simulation framework widely used in the robotics and RL community. It also offers a simplified muscle model with rigid tendons and resolves contact forces using the convex Gauss Principle. The MyoSuite~\cite{MyoSuite2022} builds on this framework, allowing for the development of high-dimensional muscle-control models which have recently gained a lot of interest from the RL community\cite{caggianoMyoDexGeneralizablePrior2023, SARBerg2023, chiappa2023latent}. Both engines achieve the required computational speed to train control policies for these high-dimensional models in under a day. See \suppref{supp:sim} for more technical details.

\begin{figure*}
    \centering
      \textcolor{ourred2}{\rule[2.5pt]{15pt}{1.5pt}} RL \textcolor{ourgray2}{\rule[2.5pt]{15pt}{1.5pt}} human-data\\
      \includegraphics[width=1.0\textwidth]{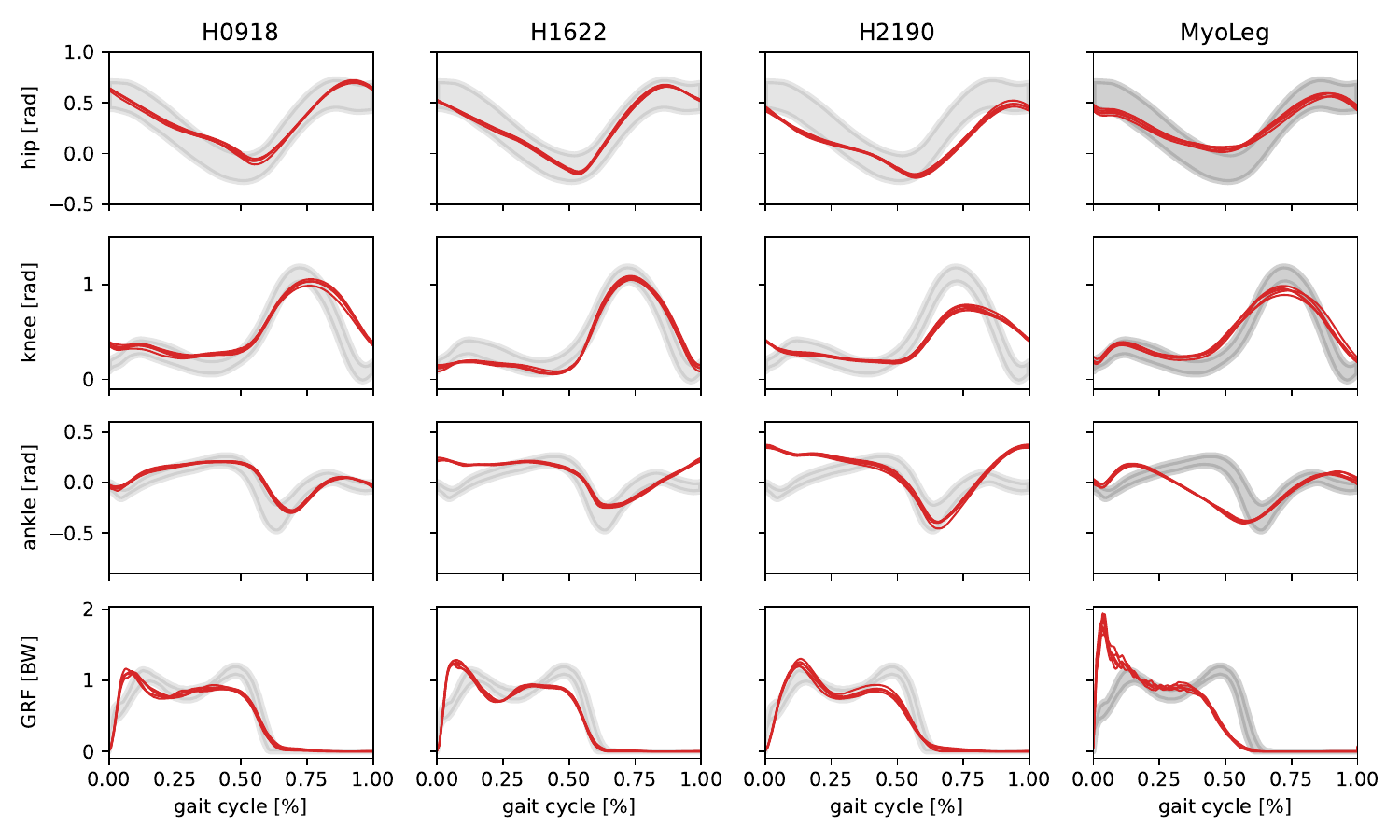}\\
    \caption{\textbf{Gait kinematics for RL agents for all models} Shown are the hip, knee, ankle and GRF values averaged over 5 rollouts of 10 s walking on flat ground. We excluded rollouts that did not achieve the whole episode length to clearly highlight the achieved kinematics. While there are slight discrepancies between experimental data (grey) and the RL behaviors (red), especially for the high-dimensional models, the proposed reward function provides a strong starting point for researchers aiming to create robust and natural controllers for high-dimensional musculoskeletal systems. Also, see the videos on the website.}
    \label{fig:rl_allmodels}
\end{figure*}
\subsection{Learned behaviors}
We first show that with our framework, we can train agents across 4 different models to produce walking gaits with the same training approach and reward function. In \Figref{fig:rl_allmodels} we compare the resulting gait kinematics against experimental data, included in the SCONE software~\cite{Geijtenbeek2019, Bovi2011}. Kinematics are shown for 5 rollouts of the most human-like policy checkpoint that was achieved over the entire training run over 10 random seeds, averaged over all gait cycles of both legs in a 10 s walk\footnote{For videos, see: \url{https://sites.google.com/view/naturalwalkingrl}}.\looseness-1

The results for the planar \planarmodel and the 3D \threedmodel model look very similar to the experimental data, even though the ankle kinematics differ slightly. While the agents achieve the most human-like gaits here, the models are also of limited complexity and applicability, compared to the high-dimensional systems, \complexmodel and \myoleg. As seen in \tabref{tab:results} and \figref{fig:rl_allmodels}, our approach still achieves periodic gaits resembling human kinematics with the difficult-to-control 80 and 90 muscle models, even though they contain more artifacts. The \complexmodel-agent exhibits less knee flexion and the \myoleg-agent lacks the double-peaked GRF structure; it also exhibits differences in the hip kinematics. Overall, the behavior of the \complexmodel model appears more natural than the one produced with the \myoleg model, see also the discussion in \secref{sec:discussion} and the supplementary videos.\looseness-1

Nevertheless, \tabref{tab:results} shows that RL gaits not only approximate human walking but are also robust and energy-efficient across all models, without changes in the reward function and only minimal changes in the hyperparameters of the RL method. We provide the training curves and additional metrics for 10 random seeds in \suppref{supp:training}.

In order to probe the robustness of our controllers, we perform roll-outs on uneven terrain, which was \textbf{not} seen during training. The entire training procedure was performed with flat ground. The generated terrain contains 10 tiles of 1~m length with random slopes of $\pm 5^{\circ}$ and is fixed for all evaluations. The behavior of the planar \planarmodel-model is compared against a popular reflex-based controller as an illustrative example, adapted from~\cite{geyer2010, Geijtenbeek2019} and included with the SCONE software. We were only able to use this simple reflex-based controller with the \planarmodel model, as it did not produce stable gaits with the other models. We train 5 reflex-based controllers with different initializations until convergence, while we use the most natural RL policy for each model and perform 20 roll-outs with randomized initial states to test the robustness. We chose this approach as reflex-based controllers are sensitive to the initial simulation state; different roll-outs would be almost identical if we would have to use similar starting states.

\begin{figure}
    \centering
   \hspace{0.35cm} \textcolor{ourred2}{\rule[2.5pt]{15pt}{1.5pt}} RL \textcolor{ourblue2}{\rule[2.5pt]{15pt}{1.5pt}} reflex-based\\
    \includegraphics[width=0.48\textwidth]{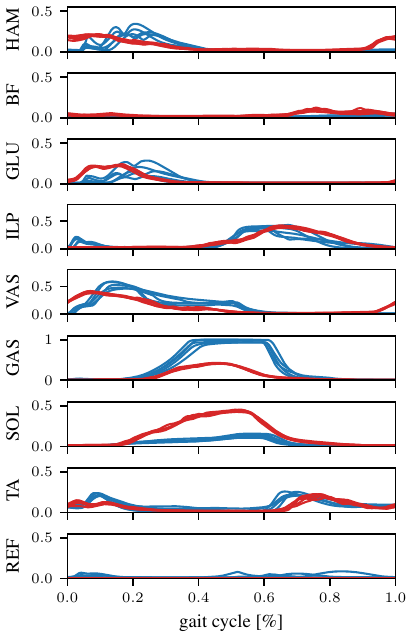}\\[-.5em]
    \caption{\textbf{Muscle activation for RL agent and reflex-based controller.} We compare muscle activities for two controller types for natural walking with the \planarmodel model. The activity for the RL agent has been clipped to $0.5$. We use 5 roll-outs of the most natural RL policy and 5 reflex-based controllers that were optimized until convergence. The initial state for the RL agent is randomized, which would cause collapse with the reflex controllers, as they are sensitive to the initial state.}
    \label{fig:activity}
\end{figure}

While both approaches adequately match human kinematics with low energy consumption in the planar case, the reflex-based controller produces more natural gaits. However, when exposed to uneven terrain, the RL agent achieves an average distance of 10.42~m, which shows that it is much more robust than the reflex controller with an average distance of 2.46~m, see \tabref{tab:results}. Both controllers also induce similar average muscle activities over the gait cycle, with the RL agent inducing less smooth activity, shown in \figref{fig:activity}. 

With the same framework, we were also able to train agents to learn maximum speed running, by simply using the achieved velocity as the velocity reward in our reward function. Additionally, the action clipping and effort costs were omitted, as energy consumption is less critical for short maximum performance tasks. See \suppref{supp:running} for these results.

As a showcase of the extreme robustness of the RL agents, we generated a difficult drawbridge-terrain task with moveable environment elements that present dynamic perturbations, see \figref{fig:running_obstacle}. We test the robustness of \threedmodel and \complexmodel RL controllers in this scenario, even though they were only ever trained on \textbf{flat} ground, and observe remarkable stability across the task. We report the data in \tabref{tab:running} and in the videos. 

Note that we tried several alternatives to our approach which yielded worse results. We performed experiments with different reward terms such as a constant instead of an adaptive effort term, with metabolic energy costs~\cite{Wang2012} or with a cost of transport\cite{cot_function,Mastrogeorgiou2023:energylegged} reward. Even though these terms sometimes lead to small muscle activity during execution, the kinematics were further away from human data. We conjecture that energy minimization is not enough of an incentive for human-like gait if the learning algorithm is as flexible as an RL agent. See also \figref{fig:ablations} for ablations of our reward function.

Larger effort term exponents, penalization of contacts between limbs or angle-based joint limit violation costs did not lead to better behavior. The prescription of hip movement at a certain frequency~(step clock), keeping certain joint angles in pre-specified positions or minimizing torso rotation helped to achieve stable gaits, but prevented effort minimization and did not lead to natural kinematics.

\begin{table}
\caption{The table shows the average cubic muscle activity~(effort), the percentage match with human experimental data~(exp. match), and the average distance walked on the rough terrain. Note that the exp. match metric measures the percentage of the gait cycle during which the trajectory perfectly lies inside the standard deviation of the experimental data. Even relatively natural gaits can still achieve a low metric if the angles are slightly shifted.}
\centering
\begin{tabular}{@{}c@{\ }ccccc@{}}
 \toprule
 controller & system & avg. effort & \multicolumn{1}{p{2em}}{experimental match} & \multicolumn{1}{p{4em}}{avg. distance~[m]}\\
 \midrule
 reflex & H0918  & 0.041 $\pm 3\times 10^{-3}$ & $0.68 \pm 0.08$ & \hspace{4pt}$2.46 \pm 0.98$\\
 RL &  H0918  & 0.013 $\pm 3\times 10^{-4}$ & $0.67 \pm 0.03$  &  $10.42 \pm 0.94$\\
 \midrule
 RL  & H1622  & 0.015 $\pm 2\times 10^{-3}$ & $0.73 \pm 0.01$& \hspace{8.1pt}$5.6 \pm 0.99$\\
 RL  & H2190  & 0.017 $\pm 1\times 10^{-5}$ & $0.50 \pm 0.01$ & $10.59 \pm 2.51$\\
 RL  & MyoLeg & 0.013 $\pm 2\times 10^{-4}$ & $0.43 \pm 0.05$ & \hspace{6pt}n.a.\\
\bottomrule
\end{tabular}
\label{tab:results}
\end{table}

\section{Discussion}
\label{sec:discussion}
As the human biomechanical system is highly redundant, there are many possible solutions to walking at a defined speed. There exists strong evidence that natural human walking is in part driven by energy-efficiency\cite{selinger2015}. Optimal control approaches have shown that natural walking kinematics can be achieved if energy optimality is considered in the cost function.\footnote{Some also suggest that muscle fatigue could be the driving factor to explain the experimentally observed kinematic patterns \cite{Ackermann2010a}.} 

However, in most RL approaches, energy consumption is either ignored, or only static action regularization is used, which affects learning but does not yield truly efficient behavior. By introducing a single reward term schedule that adapts the weighting of the energy term in the reward function depending on the current performance, we achieved energy-efficient gaits with more natural kinematics also in RL. Moreover, the adaptation algorithm (Alg. 1) and all other reward terms and their weighting coefficients are general enough to work---without any changes---across 2D and 3D models with different numbers of muscles and even different levels of biomechanical modeling accuracy. 

This is a significant step towards finding a general reward function and framework to generate natural and robust movements with RL in muscle-driven systems. Other RL frameworks that do achieve natural muscle utilization either consider low-dimensional systems~\cite{learn2run2018} or strongly rely on motion capture data~\cite{scalable2019} to render the learning problem feasible. Our approach works \textbf{without} the use of motion capture data during training and with few and very general reward terms and therefore may generalize better to other movements.

In our opinion, the only comparable work is by Weng et al.~\cite{wengnatwalking2021}. They achieved human-level kinematics on a planar human model with 18 muscles, by crafting a multi-stage learning curriculum affecting the weighting of seven reward terms. As this learning curriculum contains model-specific reward terms and adaptation procedures, we speculate that it would have to be hand-tuned for different models. 

While our approach achieved higher robustness than reflex-based controllers and kinematics closer to natural walking than previous demonstration-free RL approaches, several discrepancies to natural walking remain, see~\figref{fig:rl_allmodels} and supplementary videos. The low-dimensional models (\planarmodel and \threedmodel) in general do not present proper ankle rolling, while the high-dimensional models (\complexmodel and \myoleg) exhibit less passive leg-swing in the swing phase of the gait. 

\begin{figure}
    \centering
    \includegraphics[width=0.48\textwidth]{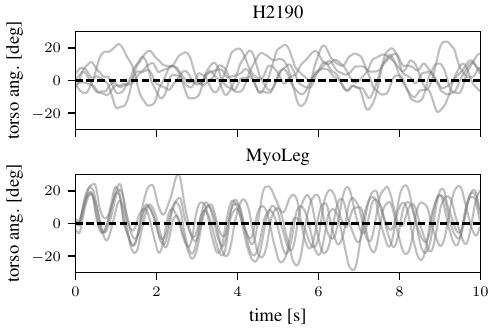}\\
    \caption{\textbf{Torso oscillations during walking.} We show the torso angle with the vertical axis for 5 rollouts of 10~s for the \complexmodel- and the \myoleg-models. The \myoleg presents stronger lateral oscillations. The dashed line shows a straight torso posture.}
    \label{fig:oscillation_walk}
\end{figure}

The behavior of the \myoleg model deviates stronger from human data than the \complexmodel, although they are similar in terms of complexity. This is most prominent in the ankle kinematics and the lack of double-peak structure in the GRFs in the \myoleg model. We also observed a tendency for unnatural lateral torso oscillations with the \myoleg model, see~\figref{fig:oscillation_walk} and the videos.

These differences in behaviors could be related to the model parametrization, as the \myoleg uses a different muscle geometry from \complexmodel and includes mesh-based contact dynamics, which might increase learning difficulty. Alternatively, the more elaborate biomechanical features in \hyfydy, such as elastic tendons~\cite{tendoncatapult2005}, non-linear foot-ground contact mechanics~\cite{reviewcontact2022}, variable pennation angles~\cite{pennationangle2016} or error-controlled integration, could account for the increased realism of the behaviors with the \hyfydy models. See \suppref{supp:sim} for more details on the simulation engines. 

Research on the contribution of biomechanical structures to the emergence of natural movement \cite{Gerritsen1998a, Haeufle2010a, John2013a, Wochner2023a} suggest that, in addition to the learning method and reward function, the biomechanical structures and modeling choices may play a crucial role in the accurate reproduction of human gait. This seems a plausible explanation for the increased realism in the \hyfydy models, as previous observations in predictive simulations suggest that e.g. an elastic tendon is beneficial for natural gait \cite{geyer2010, Wang2012, Geijtenbeek2013}.  We regard this as one interesting area of future research, which could help us better understand the fundamentals of the interaction between biomechanics and neuronal control in human locomotion.

In conclusion, we achieved highly robust walking approaching human-like kinematics and ground reaction forces. While a better degree of accuracy was achieved in simpler models, we provide first promising results for difficult-to-control 80 and 90 muscle models which are of high interest for applications in rehabilitation,
neuroscience, and computer graphics. Learning with the proposed reward function and RL framework allows for these results across several models of differing complexity and biomechanical modeling accuracy with only minimal changes in the hyperparameters of the method. We hope that this inspires researchers from both the biomechanics and the RL community to further improve on our approach and to develop tools to unravel the fundamentals of the generation of complex, robust, and energy-efficient human movement.

\section*{Acknowledgment}
Pierre Schumacher was supported by the International Max Planck Research School for Intelligent Systems (IMPRS-IS). This work was supported by the Cyber Valley Research Fund (CyVy-RF-2020-11 to DH and GM).

\bibliographystyle{IEEEtran}
\bibliography{IEEEabrv,references}

\begin{thebibliography}{10}
\providecommand{\url}[1]{#1}
\csname url@samestyle\endcsname
\providecommand{\newblock}{\relax}
\providecommand{\bibinfo}[2]{#2}
\providecommand{\BIBentrySTDinterwordspacing}{\spaceskip=0pt\relax}
\providecommand{\BIBentryALTinterwordstretchfactor}{4}
\providecommand{\BIBentryALTinterwordspacing}{\spaceskip=\fontdimen2\font plus
\BIBentryALTinterwordstretchfactor\fontdimen3\font minus
  \fontdimen4\font\relax}
\providecommand{\BIBforeignlanguage}[2]{{%
\expandafter\ifx\csname l@#1\endcsname\relax
\typeout{** WARNING: IEEEtran.bst: No hyphenation pattern has been}%
\typeout{** loaded for the language `#1'. Using the pattern for}%
\typeout{** the default language instead.}%
\else
\language=\csname l@#1\endcsname
\fi
#2}}
\providecommand{\BIBdecl}{\relax}
\BIBdecl

\bibitem{inversecostfunctiontuning2011}
\BIBentryALTinterwordspacing
B.~Berret, E.~Chiovetto, F.~Nori, and T.~Pozzo, ``Evidence for composite cost
  functions in arm movement planning: An inverse optimal control approach,''
  \emph{PLOS Computational Biology}, vol.~7, no.~10, pp. 1--18, 10 2011.
  [Online]. Available: \url{https://doi.org/10.1371/journal.pcbi.1002183}
\BIBentrySTDinterwordspacing

\bibitem{amp2021}
\BIBentryALTinterwordspacing
X.~B. Peng, Z.~Ma, P.~Abbeel, S.~Levine, and A.~Kanazawa, ``Amp: Adversarial
  motion priors for stylized physics-based character control,'' \emph{ACM
  Trans. Graph.}, vol.~40, no.~4, Jul. 2021. [Online]. Available:
  \url{http://doi.acm.org/10.1145/3450626.3459670}
\BIBentrySTDinterwordspacing

\bibitem{costfunctionvaries2023}
J.~Weng, E.~Hashemi, and A.~Arami, ``Human gait cost function varies with
  walking speed: An inverse optimal control study,'' \emph{IEEE Robotics and
  Automation Letters}, vol.~8, no.~8, pp. 4777--4784, 2023.

\bibitem{wengnatwalking2021}
------, ``Natural walking with musculoskeletal models using deep reinforcement
  learning,'' \emph{IEEE Robotics and Automation Letters}, vol.~6, no.~2, pp.
  4156--4162, 2021.

\bibitem{gaitnet2022}
\BIBentryALTinterwordspacing
J.~Park, S.~Min, P.~S. Chang, J.~Lee, M.~S. Park, and J.~Lee, ``Generative
  gaitnet,'' in \emph{ACM SIGGRAPH 2022 Conference Proceedings}, ser. SIGGRAPH
  '22.\hskip 1em plus 0.5em minus 0.4em\relax New York, NY, USA: Association
  for Computing Machinery, 2022. [Online]. Available:
  \url{https://doi.org/10.1145/3528233.3530717}
\BIBentrySTDinterwordspacing

\bibitem{accelerated2022}
\BIBentryALTinterwordspacing
J.~Xu, M.~Macklin, V.~Makoviychuk, Y.~Narang, A.~Garg, F.~Ramos, and
  W.~Matusik, ``Accelerated policy learning with parallel differentiable
  simulation,'' in \emph{International Conference on Learning Representations},
  2022. [Online]. Available: \url{https://openreview.net/forum?id=ZSKRQMvttc}
\BIBentrySTDinterwordspacing

\bibitem{schumacher2023deprl}
\BIBentryALTinterwordspacing
P.~Schumacher, D.~Haeufle, D.~B{\"u}chler, S.~Schmitt, and G.~Martius,
  ``{DEP}-{RL}: Embodied exploration for reinforcement learning in overactuated
  and musculoskeletal systems,'' in \emph{The Eleventh International Conference
  on Learning Representations}, 2023. [Online]. Available:
  \url{https://openreview.net/forum?id=C-xa_D3oTj6}
\BIBentrySTDinterwordspacing

\bibitem{Wang2012}
J.~Wang, S.~Hamner, S.~Delp, and V.~Koltun, ``{Optimizing Locomotion
  Controllers Using Biologically-Based Actuators and Objectives},'' \emph{ACM
  Trans. on Graphics}, vol.~31, no.~4, p.~25, 2012.

\bibitem{learn2run2018}
{\L}.~Kidzi{\'{n}}ski, S.~P. Mohanty, C.~F. Ong, Z.~Huang, S.~Zhou,
  A.~Pechenko, A.~Stelmaszczyk, P.~Jarosik, M.~Pavlov, S.~Kolesnikov, S.~Plis,
  Z.~Chen, Z.~Zhang, J.~Chen, J.~Shi, Z.~Zheng, C.~Yuan, Z.~Lin,
  H.~Michalewski, P.~Milos, B.~Osinski, A.~Melnik, M.~Schilling, H.~Ritter,
  S.~F. Carroll, J.~Hicks, S.~Levine, M.~Salath{\'e}, and S.~Delp, ``Learning
  to run challenge solutions: Adapting reinforcement learning methods for
  neuromusculoskeletal environments,'' in \emph{The NIPS '17 Competition:
  Building Intelligent Systems}, S.~Escalera and M.~Weimer, Eds.\hskip 1em plus
  0.5em minus 0.4em\relax Cham: Springer International Publishing, 2018, pp.
  121--153.

\bibitem{songDeepReinforcementLearning2020a}
\BIBentryALTinterwordspacing
S.~Song, {\L}.~Kidzi{\'{n}}ski, X.~B. Peng, C.~Ong, J.~Hicks, S.~Levine, C.~G.
  Atkeson, and S.~L. Delp, ``Deep reinforcement learning for modeling human
  locomotion control in neuromechanical simulation,'' \emph{Journal of
  NeuroEngineering and Rehabilitation}, vol.~18, no.~1, p. 126, Aug 2021.
  [Online]. Available: \url{https://doi.org/10.1186/s12984-021-00919-y}
\BIBentrySTDinterwordspacing

\bibitem{SARBerg2023}
C.~Berg, V.~Caggiano, and V.~Kumar, ``Sar: Generalization of physiological
  agility and dexterity via synergistic action representation,'' 2023.

\bibitem{geyer2010}
H.~Geyer and H.~Herr, ``{{A} muscle-reflex model that encodes principles of
  legged mechanics produces human walking dynamics and muscle activities},''
  \emph{IEEE Trans Neural Syst Rehabil Eng}, vol.~18, no.~3, pp. 263--273, Jun
  2010.

\bibitem{Song2013}
S.~Song and H.~Geyer, ``Generalization of a muscle-reflex control model to {3D}
  walking,'' in \emph{2013 35th Annual International Conference of the {IEEE}
  Engineering in Medicine and Biology Society ({EMBC})}.\hskip 1em plus 0.5em
  minus 0.4em\relax IEEE, Jul. 2013.

\bibitem{Geijtenbeek2013}
T.~Geijtenbeek, M.~van~de Panne, and A.~F. van~der Stappen, ``Flexible
  muscle-based locomotion for bipedal creatures,'' \emph{ACM Transactions on
  Graphics}, vol.~32, no.~6, 2013.

\bibitem{Song2017a}
S.~Song and H.~Geyer, ``{Evaluation of a Neuromechanical Walking Control Model
  Using Disturbance Experiments},'' \emph{Frontiers in Computational
  Neuroscience}, vol.~11, no.~3, 2017.

\bibitem{Haeufle2018a}
D.~F.~B. Haeufle, B.~Schmortte, H.~Geyer, R.~M{\"{u}}ller, and .~Schmitt, Syn,
  ``{The Benefit of Combining Neuronal Feedback and Feed-Forward Control for
  Robustness in Step Down Perturbations of Simulated Human Walking Depends on
  the Muscle Function},'' \emph{Frontiers in Computational Neuroscience},
  vol.~12, no.~80, 2018.

\bibitem{Ramadan2022a}
R.~Ramadan, H.~Geyer, J.~Jeka, G.~Schöner, and H.~Reimann, ``A neuromuscular
  model of human locomotion combines spinal reflex circuits with voluntary
  movements,'' \emph{Scientific Reports}, vol.~12, no.~1, may 2022.

\bibitem{Schreff2022a}
L.~Schreff, D.~F.~B. Haeufle, J.~Vielemeyer, and R.~Müller, ``Evaluating
  anticipatory control strategies for their capability to cope with step-down
  perturbations in computer simulations of human walking,'' \emph{Scientific
  Reports}, vol.~12, no.~1, jun 2022.

\bibitem{efficientspeed2007}
B.~J. Mohler, W.~B. Thompson, S.~H. Creem-Regehr, H.~L. Pick, Jr, and W.~H.
  Warren, Jr, ``\BIBforeignlanguage{en}{Visual flow influences gait transition
  speed and preferred walking speed},'' \emph{\BIBforeignlanguage{en}{Exp.
  Brain Res.}}, vol. 181, no.~2, pp. 221--228, Aug. 2007.

\bibitem{rudin2022skills}
N.~Rudin, D.~Hoeller, M.~Bjelonic, and M.~Hutter, ``Advanced skills by learning
  locomotion and local navigation end-to-end,'' in \emph{2022 IEEE/RSJ
  International Conference on Intelligent Robots and Systems (IROS)}, 2022, pp.
  2497--2503.

\bibitem{dajiroenergyefficient2015}
\BIBentryALTinterwordspacing
D.~Abe, Y.~Fukuoka, and M.~Horiuchi, ``Economical speed and energetically
  optimal transition speed evaluated by gross and net oxygen cost of transport
  at different gradients,'' \emph{PLOS ONE}, vol.~10, no.~9, pp. 1--14, 09
  2015. [Online]. Available: \url{https://doi.org/10.1371/journal.pone.0138154}
\BIBentrySTDinterwordspacing

\bibitem{raffaltenergyefficient2017}
\BIBentryALTinterwordspacing
P.~C. Raffalt, M.~K. Guul, A.~N. Nielsen, S.~Puthusserypady, and T.~Alkj{\ae}r,
  ``Economy, movement dynamics, and muscle activity of human walking at
  different speeds,'' \emph{Scientific Reports}, vol.~7, no.~1, p. 43986, Mar
  2017. [Online]. Available: \url{https://doi.org/10.1038/srep43986}
\BIBentrySTDinterwordspacing

\bibitem{ackermann2010}
\BIBentryALTinterwordspacing
M.~Ackermann and A.~J. {van den Bogert}, ``Optimality principles for
  model-based prediction of human gait,'' \emph{Journal of Biomechanics},
  vol.~43, no.~6, pp. 1055--1060, 2010. [Online]. Available:
  \url{https://www.sciencedirect.com/science/article/pii/S0021929009007210}
\BIBentrySTDinterwordspacing

\bibitem{zahavy2023discovering}
\BIBentryALTinterwordspacing
T.~Zahavy, Y.~Schroecker, F.~Behbahani, K.~Baumli, S.~Flennerhag, S.~Hou, and
  S.~Singh, ``Discovering policies with {DOM}i{NO}: Diversity optimization
  maintaining near optimality,'' in \emph{The Eleventh International Conference
  on Learning Representations}, 2023. [Online]. Available:
  \url{https://openreview.net/forum?id=kjkdzBW3b8p}
\BIBentrySTDinterwordspacing

\bibitem{2021pebble}
K.~Lee, L.~Smith, and P.~Abbeel, ``Pebble: Feedback-efficient interactive
  reinforcement learning via relabeling experience and unsupervised
  pre-training,'' in \emph{International Conference on Machine Learning}, 2021.

\bibitem{grfwalking1989}
\BIBentryALTinterwordspacing
J.~NILSSON and A.~THORSTENSSON, ``Ground reaction forces at different speeds of
  human walking and running,'' \emph{Acta Physiologica Scandinavica}, vol. 136,
  no.~2, pp. 217--227, 1989. [Online]. Available:
  \url{https://onlinelibrary.wiley.com/doi/abs/10.1111/j.1748-1716.1989.tb08655.x}
\BIBentrySTDinterwordspacing

\bibitem{Geijtenbeek2019}
\BIBentryALTinterwordspacing
T.~Geijtenbeek, ``{SCONE: Open Source Software for Predictive Simulation of
  Biological Motion},'' \emph{Journal of Open Source Software}, vol.~4, no.~38,
  p. 1421, 2019. [Online]. Available: \url{https://scone.software}
\BIBentrySTDinterwordspacing

\bibitem{Delp1990}
S.~L. Delp, J.~P. Loan, M.~G. Hoy, and F.~E. Zajac, ``{An interactive
  Graphics-Based Model of the Lower Extremity to Study Orthopaedic Surgical
  Procedures},'' pp. 757 -- 767, 1990.

\bibitem{Rajagopal2016}
A.~Rajagopal, C.~Dembia, M.~DeMers, D.~Delp, J.~Hicks, and S.~Delp, ``{Full
  body musculoskeletal model for muscle-driven simulation of human gait},''
  \emph{IEEE Transactions on Biomedical Engineering}, vol.~63, no.~10, pp.
  2068--2079, 2016.

\bibitem{Christophy2012}
M.~Christophy, N.~A. {Faruk Senan}, J.~C. Lotz, and O.~M. O'Reilly, ``{A
  musculoskeletal model for the lumbar spine.}'' \emph{Biomechanics and
  modeling in mechanobiology}, vol.~11, no. 1-2, pp. 19--34, jan 2012.

\bibitem{Geijtenbeek2021}
\BIBentryALTinterwordspacing
T.~Geijtenbeek, ``The {Hyfydy} simulation software,'' 11 2021. [Online].
  Available: \url{https://hyfydy.com}
\BIBentrySTDinterwordspacing

\bibitem{MyoSuite2022}
\BIBentryALTinterwordspacing
V.~Caggiano, H.~Wang, G.~Durandau, M.~Sartori, and V.~Kumar, ``Myosuite -- a
  contact-rich simulation suite for musculoskeletal motor control,''
  \url{https://github.com/facebookresearch/myosuite}, 2022. [Online].
  Available: \url{https://sites.google.com/view/myosuite}
\BIBentrySTDinterwordspacing

\bibitem{Seth2018}
A.~Seth \emph{et~al.}, ``Opensim: Simulating musculoskeletal dynamics and
  neuromuscular control to study human and animal movement,'' \emph{PLoS
  computational biology}, vol.~14, no.~7, p. e1006223, 2018.

\bibitem{caggianoMyoDexGeneralizablePrior2023}
\BIBentryALTinterwordspacing
V.~Caggiano, S.~Dasari, and V.~Kumar. {{MyoDex}}: {{A Generalizable Prior}} for
  {{Dexterous Manipulation}}. [Online]. Available:
  \url{https://openreview.net/forum?id=iYBTiYzN0A}
\BIBentrySTDinterwordspacing

\bibitem{chiappa2023latent}
A.~S. Chiappa, A.~M. Vargas, A.~Z. Huang, and A.~Mathis, ``Latent exploration
  for reinforcement learning,'' 2023.

\bibitem{Bovi2011}
G.~Bovi, M.~Rabuffetti, P.~Mazzoleni, and M.~Ferrarin, ``{A multiple-task gait
  analysis approach: Kinematic, kinetic and EMG reference data for healthy
  young and adult subjects},'' \emph{Gait \& Posture}, vol.~33, no.~1, pp.
  6--13, jan 2011.

\bibitem{cot_function}
\BIBentryALTinterwordspacing
K.~Veerkamp, N.~Waterval, T.~Geijtenbeek, C.~Carty, D.~Lloyd, J.~Harlaar, and
  M.~{van der Krogt}, ``Evaluating cost function criteria in predicting healthy
  gait,'' \emph{Journal of Biomechanics}, vol. 123, p. 110530, 2021. [Online].
  Available:
  \url{https://www.sciencedirect.com/science/article/pii/S0021929021003110}
\BIBentrySTDinterwordspacing

\bibitem{Mastrogeorgiou2023:energylegged}
A.~Mastrogeorgiou, A.~Papatheodorou, K.~Koutsoukis, and E.~Papadopoulos,
  ``Learning energy-efficient trotting for legged robots,'' in \emph{Robotics
  in Natural Settings}.\hskip 1em plus 0.5em minus 0.4em\relax Cham: Springer
  International Publishing, 2023, pp. 204--215.

\bibitem{selinger2015}
\BIBentryALTinterwordspacing
J.~Selinger, S.~O’Connor, J.~Wong, and J.~Donelan, ``Humans can continuously
  optimize energetic cost during walking,'' \emph{Current Biology}, vol.~25,
  no.~18, pp. 2452--2456, 2015. [Online]. Available:
  \url{https://www.sciencedirect.com/science/article/pii/S0960982215009586}
\BIBentrySTDinterwordspacing

\bibitem{Ackermann2010a}
M.~Ackermann and A.~J. van~den Bogert, ``Optimality principles for model-based
  prediction of human gait,'' \emph{Journal of Biomechanics}, vol.~43, no.~6,
  pp. 1055--1060, apr 2010.

\bibitem{scalable2019}
\BIBentryALTinterwordspacing
S.~Lee, M.~Park, K.~Lee, and J.~Lee, ``Scalable muscle-actuated human
  simulation and control,'' \emph{ACM Trans. Graph.}, vol.~38, no.~4, jul 2019.
  [Online]. Available: \url{https://doi.org/10.1145/3306346.3322972}
\BIBentrySTDinterwordspacing

\bibitem{tendoncatapult2005}
\BIBentryALTinterwordspacing
M.~Ishikawa, P.~V. Komi, M.~J. Grey, V.~Lepola, and G.-P. Bruggemann,
  ``Muscle-tendon interaction and elastic energy usage in human walking,''
  \emph{Journal of Applied Physiology}, vol.~99, no.~2, pp. 603--608, 2005,
  pMID: 15845776. [Online]. Available:
  \url{https://doi.org/10.1152/japplphysiol.00189.2005}
\BIBentrySTDinterwordspacing

\bibitem{reviewcontact2022}
\BIBentryALTinterwordspacing
L.~Saraiva, M.~{Rodrigues da Silva}, F.~Marques, M.~{Tavares da Silva}, and
  P.~Flores, ``A review on foot-ground contact modeling strategies for human
  motion analysis,'' \emph{Mechanism and Machine Theory}, vol. 177, p. 105046,
  2022. [Online]. Available:
  \url{https://www.sciencedirect.com/science/article/pii/S0094114X22002932}
\BIBentrySTDinterwordspacing

\bibitem{pennationangle2016}
R.~Sopher, A.~Amis, D.~Davies, and J.~Jeffers, ``The influence of muscle
  pennation angle and cross-sectional area on contact forces in the ankle
  joint,'' \emph{The Journal of Strain Analysis for Engineering Design},
  vol.~52, 09 2016.

\bibitem{Gerritsen1998a}
K.~G. Gerritsen, A.~J. van~den Bogert, M.~Hulliger, and .~Zernicke, Ronald~F,
  ``{Intrinsic Muscle Properties Facilitate Locomotor Control - A Computer
  Simulation Study},'' \emph{Motor Control}, vol.~2, no.~3, pp. 206--220, jul
  1998.

\bibitem{Haeufle2010a}
D.~F. Haeufle, S.~Grimmer, and .~Seyfarth, A., ``{The role of intrinsic muscle
  properties for stable hopping - stability is achieved by the force-velocity
  relation},'' \emph{Bioinspiration {\&} Biomimetics}, vol.~5, no.~1, p.
  016004, 2010.

\bibitem{John2013a}
C.~T. John, F.~C. Anderson, J.~S. Higginson, and S.~L. Delp, ``{Stabilisation
  of walking by intrinsic muscle properties revealed in a three-dimensional
  muscle-driven simulation.}'' \emph{Computer methods in biomechanics and
  biomedical engineering}, vol.~16, no.~4, pp. 451--62, apr 2013.

\bibitem{Wochner2023a}
I.~Wochner, P.~Schumacher, G.~Martius, D.~B\"uchler, S.~Schmitt, and
  D.~Haeufle, ``Learning with muscles: Benefits for data-efficiency and
  robustness in anthropomorphic tasks,'' in \emph{Proceedings of The 6th
  Conference on Robot Learning}, ser. Proceedings of Machine Learning Research,
  K.~Liu, D.~Kulic, and J.~Ichnowski, Eds., vol. 205.\hskip 1em plus 0.5em
  minus 0.4em\relax PMLR, 14--18 Dec 2023, pp. 1178--1188.

\bibitem{Millard2013}
M.~Millard, T.~Uchida, A.~Seth, and S.~L. Delp, ``{Flexing computational
  muscle: modeling and simulation of musculotendon dynamics.}'' \emph{Journal
  of biomechanical engineering}, vol. 135, no.~2, p. 021005, feb 2013.

\bibitem{Hunt1975}
K.~H. Hunt and F.~R.~E. Crossley, ``{Coefficient of Restitution Interpreted as
  Damping in Vibroimpact},'' \emph{Journal of Applied Mechanics}, vol.~42,
  no.~2, p. 440, jun 1975.

\bibitem{Sherman2011}
M.~a. Sherman, A.~Seth, and S.~L. Delp, ``{Simbody: multibody dynamics for
  biomedical research},'' \emph{Procedia IUTAM}, vol.~2, pp. 241--261, jan
  2011.

\bibitem{hamner2013}
\BIBentryALTinterwordspacing
S.~R. Hamner and S.~L. Delp, ``Muscle contributions to fore-aft and vertical
  body mass center accelerations over a range of running speeds,''
  \emph{Journal of Biomechanics}, vol.~46, no.~4, pp. 780--787, 2013. [Online].
  Available:
  \url{https://www.sciencedirect.com/science/article/pii/S0021929012006768}
\BIBentrySTDinterwordspacing

\bibitem{pardo2020tonic}
F.~Pardo, ``Tonic: A deep reinforcement learning library for fast prototyping
  and benchmarking,'' \emph{arXiv preprint arXiv:2011.07537}, 2020.

\end{thebibliography}
\appendix

\subsection{Simulation Engines}
\label{supp:sim}
 The \hyfydy and \mujoco simulation engines differ in these key areas:
\myparagraph{Musculotendon dynamics} The muscle model in \hyfydy is based on Millard at el. \cite{Millard2013} and includes elastic tendons, muscle pennation, and muscle fiber damping. The \mujoco muscle model is based on a simplified Hill-type model, parameterized to match existing OpenSim models~\cite{MyoSuite2022}, and supports only rigid tendons and does not include variable pennation angles.
\myparagraph{Contact forces} \hyfydy uses the Hunt-Crossly \cite{Hunt1975} contact model with non-linear damping to generate contact forces, with a friction cone based on dynamic, static, and viscous friction coefficients \cite{Sherman2011}. \mujoco contacts are rigid, with a friction pyramid instead of a cone, and without separate coefficients for dynamic and viscous friction.
\myparagraph{Contact geometry} The \mujoco model uses a convex mesh for foot geometry, while in the \hyfydy models the foot geometry is approximated using three contact spheres.
\myparagraph{Integration step size} \hyfydy uses an error-controlled integrator with variable step size, while \mujoco uses a fixed step size and no error control. The average simulation step size in \hyfydy is around 0.00014s (7000\,Hz) for the H2190 model, compared to the fixed MyoSuite step size of 0.001s (1000\,Hz) for the \myoleg model.
\subsection{Training Curves}
\label{supp:training}
\begin{figure}
    \centering
    \hspace{1.2cm}\textcolor{ourblue2}{\rule[2.5pt]{15pt}{1.5pt}} H0918 \textcolor{ourorange2}{\rule[2.5pt]{15pt}{1.5pt}} H1622 \textcolor{ourgreen2}{\rule[2.5pt]{15pt}{1.5pt}} H2190 \textcolor{ourred2}{\rule[2.5pt]{15pt}{1.5pt}} MyoLeg\\
    \vspace{0.1cm}
    \includegraphics[width=0.48\textwidth]{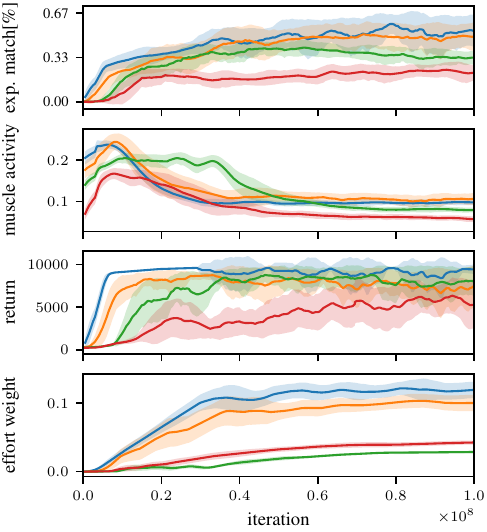}
    \caption{\textbf{Training curves for the walking task.} We present the evolution of the match with experimental data (exp. match), the averaged muscle activity, the task performance, and the effort cost weight. The cost weight increases more slowly for the more complex models, showing its adaptive nature.}
    \label{supp:trainingcurve}
\end{figure}

Here we present more detailed results about the training evolution in \figref{supp:trainingcurve}. We plot the experimental match percentage between the collected gait-cycle averaged data and experimental human data, the muscle-averaged effort, the training returns, and the weight that the effort-reward term has over training. This weight is adapted over time and depends on the agent's performance. It increases slower for the complex models and saturates at smaller values. It can also be seen that the returns for the \myoleg are generally smaller than for the other models. We observed that there was more variance over training and over different seeds for the \myoleg-agent, leading to much smaller averaged returns. It was still possible to find a training checkpoint that achieved robust, close-to human-like walking for this model. 
\subsection{Running}
\label{supp:running}
We performed maximum-speed running experiments with every model. While most reward terms remained identical to the natural walking case, we replace the external task reward by the velocity of the center of mass $r_{\mathrm{vel}} = v$ and removed energetic constraints such as the muscle excitation clipping and the effort cost term. The gait-cycle- and leg-averaged kinematics are shown in \figref{supp:running}. As this task is a maximum performance movement, we have equalized the forces between the \hyfydy- and \mujoco-based models, as the \hyfydy-models in the main experiments are generally based on experimental data with weaker maximum isometric muscle forces~\cite{Delp1990}. Note that we added a negative reward for self-collision forces for the running tasks, as the agents would often cross their legs and hit them against each other, thereby hopping instead of running. 

\begin{figure*}
\begin{subfigure}[b]{0.5\textwidth}
    \centering
    \includegraphics[width=1.0\textwidth]{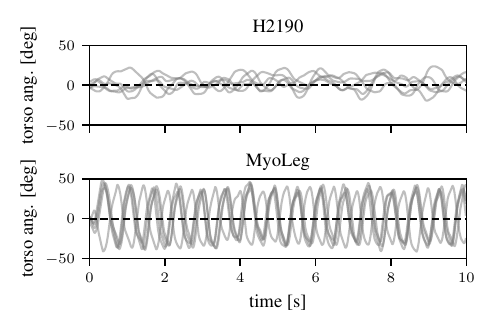}\\
    \vspace{-10pt}
    \caption{\textbf{Torso oscillations during flat-ground running.} We show the torso angle with the vertical axis for 5 rollouts of 10~s for the \complexmodel and the \myoleg models. The \myoleg presents strong lateral oscillations. The dashed line shows a straight torso posture. {\color{white}textesttttttttttttttttttttt\\ttttsssssssssssssssssssssss}}
    \label{supp:oscillation_run}
    \end{subfigure}
    \begin{subfigure}[b]{0.5\textwidth}
    \centering
   \includegraphics[width=0.7\textwidth]{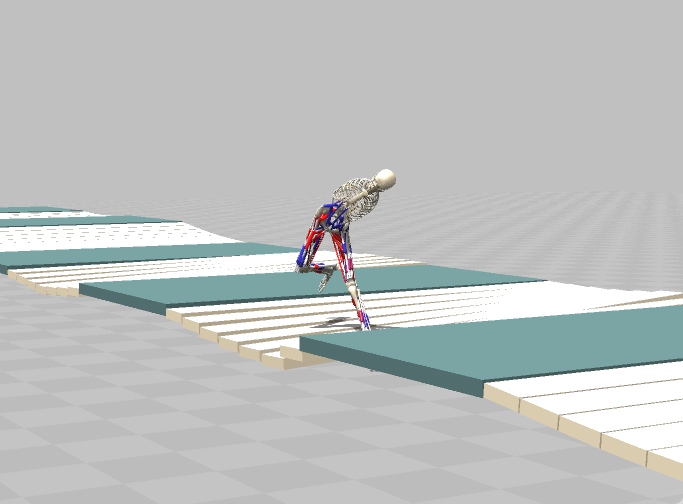}\\
   \vspace{5pt}
    \caption{\textbf{Dynamic terrain for running.} We probe the robustness of our policies trained for the \threedmodel and the \complexmodel with challenging obstacles. The tiles of the bridge rotate around the central axis and hang downwards, similar to a drawbridge. The agents were trained on \textbf{flat} ground and only have access to proprioceptive feedback.}
    \label{fig:running_obstacle}
    \end{subfigure}
    \caption{Torso oscillations for running and dynamic perturbation environment.}
\end{figure*}

Even though there remains a stronger discrepancy between the produced kinematics and the experimental data than for walking, the hip movement and GRFs are generally well aligned for the \hyfydy-models. The \myoleg-model presents very strong lateral torso oscillations during running, see also \figref{supp:oscillation_run}. In future work, biological objectives such as head stabilization or the inclusion of arms in the model might minimize some of these artifacts. See \tabref{tab:running} for the maximum running velocities for each model. 

We also performed robustness experiments on a challenging obstacle course, see \figref{fig:running_obstacle} and supplementary videos.

\begin{table}
\caption{Maximum running velocity for different models, expressed in \nicefrac{m}{s} and total achieved distance in the rough terrain environment. We only show the maximum speed over 20 roll-outs for each model to show the largest velocity that we were able to achieve. For the rough terrain, we also record 20 roll-outs for each agent. We do not perform this experiment for the \planarmodel model, as the 3D nature of the terrain is not applicable to it, and we do not apply it to the \myoleg model, as the terrain has not been implemented in the \mujoco simulator.}
\centering
\begin{tabular}{ccccc}
 \toprule 
 system & H0918 & H1622 & H2190 & 
 MyoLeg\\
 
 \midrule
 max. velocity & 5.38 & 5.04 & 6.49 & 5.44\\ 
 achieved distance & n.a. & $9.87 \pm 4.27$ & $10.45 \pm 4.77$ & n.a.\\
\bottomrule
\end{tabular}
\label{tab:running}
\end{table}

\begin{figure*}
    \centering
      \textcolor{ourred2}{\rule[2.5pt]{15pt}{1.5pt}} RL \,\textcolor{ourgray2}{\rule[2.5pt]{15pt}{1.5pt}} human-data\\
    \includegraphics[width=1.0\textwidth]{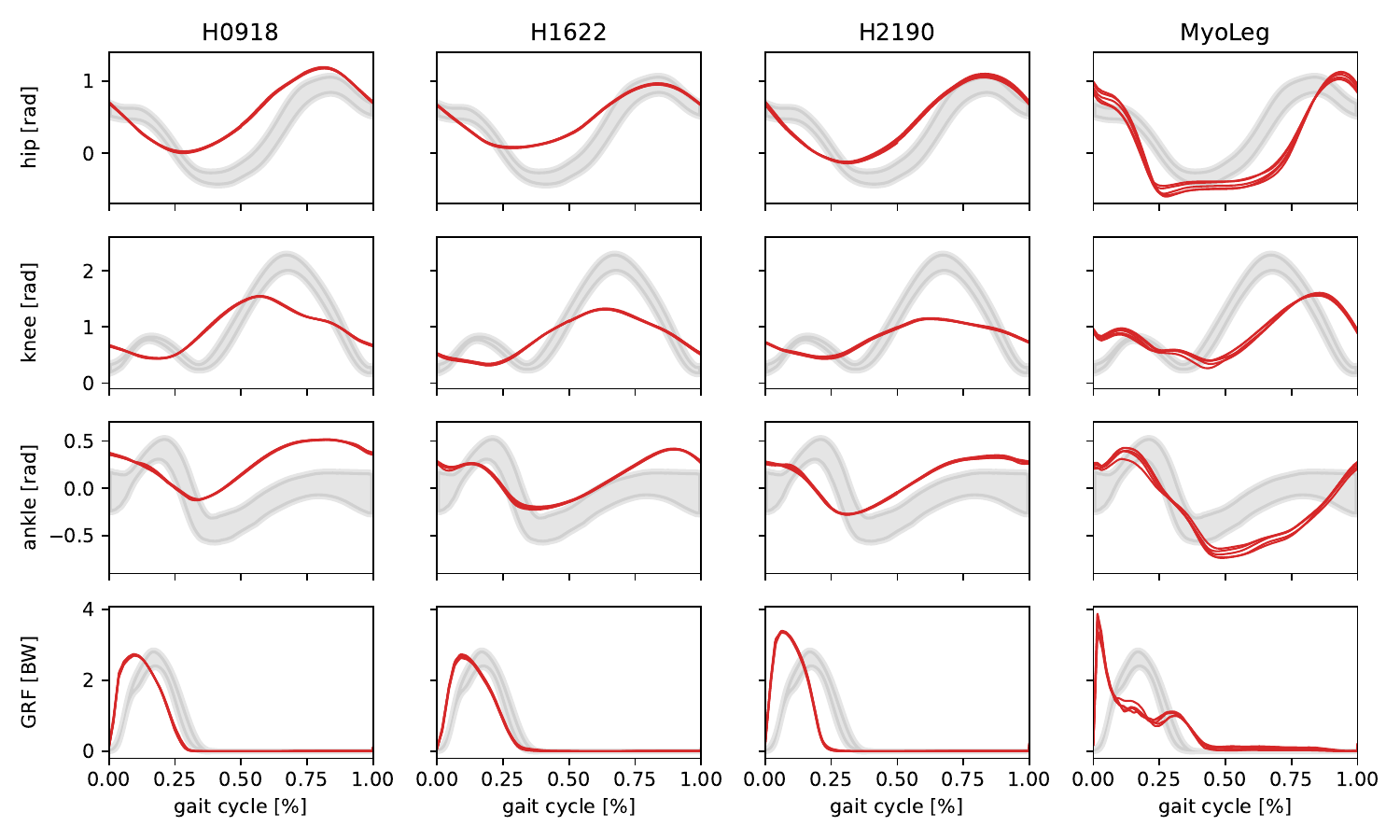}\\
    \caption{\textbf{Gait-cycle kinematics for running.} The experimental data shows human subjects running at $5 $~\nicefrac{m}{s} and was extracted from \cite{hamner2013}.}
    \label{fig:running}
\end{figure*}
\subsection{Reward function ablation}
\begin{figure*}
    \centering
          \hspace{40pt}\textcolor{ourred2}{\rule[2.5pt]{15pt}{1.5pt}} ours \,\textcolor{ourblue2}{\rule[2.5pt]{15pt}{1.5pt}} no-adapt \,\textcolor{ourorange2}{\rule[2.5pt]{15pt}{1.5pt}} no-effort \,\textcolor{ourgreen2}{\rule[2.5pt]{15pt}{1.5pt}} only-vel\\
          \vspace{5pt}

    \includegraphics[width=0.9\textwidth]{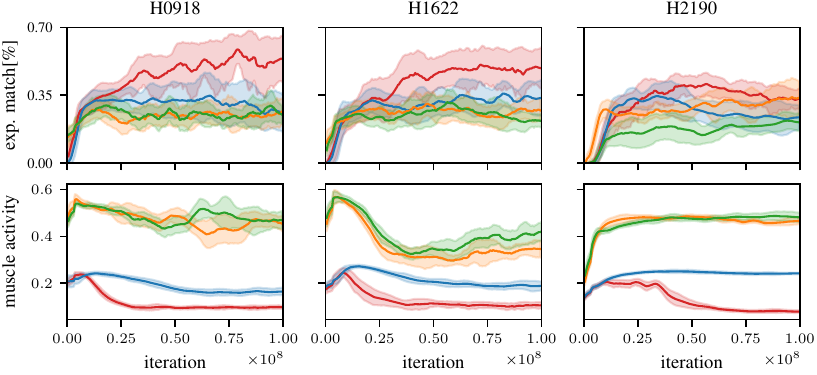}\\
    \caption{\textbf{Cost function ablations.} We show several ablations of our cost function and plot the average match with experimental human data, as detailed in the main paper, as well as the average muscle activity. A natural gait is generally characterized by a large experimental match as well as minimal muscle activity. Different ablations are shown: The adaptive effort term is zero ($\alpha(t) = 0$): no-adapt. The entire effort cost term is zero ($c_{\mathrm{effort}}=0$) and we deactivate the action clipping: no-effort. We only reward with the velocity reward term ($c_\mathrm{effort} = 0$ \& $c_{\mathrm{pain}} =0$): only-vel. Only the combined cost function achieves a close resemblance to natural gait with low muscle activity. Leaving out the pain-related costs leads to the worst gait trajectories, while a combination of the effort cost terms and the adaptive cost term is needed to achieve the lowest muscle activity.}
    \label{fig:ablations}
\end{figure*}

We perform ablations on our reward, see \figref{fig:ablations}. Throughout all considered variations, only the full reward functions leads to gaits resembling human kinematics with low muscle activity across all models.

\subsection{Hyperparameters}
Used hyperparameter settings for the RL agent, DEP and the cost function are shown in \tabref{supp:params}. Non-reported RL values are left to their default setting in TonicRL~\cite{pardo2020tonic}. See \cite{schumacher2023deprl} for an explanation of the DEP-specific terms. The RL parameters were held constant, except for an increase in network capacity for \complexmodel and \myoleg.
\begin{table*}
    \centering
     \caption{Hyperparameters for all algorithms.}
    \label{supp:params}
     \begin{subtable}[t]{.5\textwidth}
        \centering
        \caption{DEP settings.}
        \begin{tabular}{@{}lll@{}}
        \toprule
        &\textbf{Parameter} & \textbf{Value} \\

        \midrule
        DEP & $\kappa$ & 1200 \\
         & $\tau$ & 40 \\
         & buffer size & 200\\
         & bias rate & 0.002 \\
         & s4avg & $2$ \\
         & time dist ($\Delta t$) & $5$\\
         \midrule
        integration & $p_{\mathrm{switch}}$ & $3.71\times 10^{-4}$ \\
        & $H_{\mathrm{DEP}}$ & 8 \\
        & test episode & 3 \\
        & force scale & n.a. \\
        \bottomrule
        \end{tabular}
    \end{subtable}%
 \begin{subtable}[t]{.5\textwidth}
        \centering
        \caption{MPO settings}
          \begin{tabular}{@{}lll@{}}
        \toprule
        &\textbf{Parameter} & \textbf{Value} \\
        \midrule
                & buffer size  & 1e6 \\
        & batch size & 256 \\
        & steps before batches & 2e5 \\
        & steps between batches & 1000 \\
        & number of batches & 30 \\
        & n-step return & 1 \\
        & n parallel & 20 \\
        & n sequential & 10 \\
        & hidden layers & 2 \\
        & hidden sizes  & 256 \\
        & lr$_{\mathrm{actor}}$ & $3\times 10^{-4}$ \\
        & lr$_{\mathrm{critic}}$ & $3\times 10^{-4}$ \\
        & lr$_{\mathrm{dual}}$ & $1\times 10^{-2}$ \\

        \bottomrule
        \end{tabular}
    \end{subtable}\\
    \vspace{0.3cm}
        \begin{subtable}[t]{.5\textwidth}
        \centering
        \caption{MPO setting changes for \complexmodel and \myoleg.}
          \begin{tabular}{@{}lll@{}}
        \toprule
        &\textbf{Parameter} & \textbf{Value} \\
        \midrule
                & hidden sizes  & 1024 \\
        & lr$_{\mathrm{actor}}$ & $3.53\times 10^{-5}$ \\
        & lr$_{\mathrm{critic}}$ & $6.08\times 10^{-5}$ \\
        & lr$_{\mathrm{dual}}$ & $2.13\times 10^{-3}$ \\
        \bottomrule
        \end{tabular}
        \end{subtable}%
 \begin{subtable}[t]{.5\textwidth}
        \centering
        \caption{Cost function settings.}
        \begin{tabular}{@{}llll@{}}
        \toprule
        &\textbf{Parameter} & \textbf{Value} & \textbf{Meaning}\\
        \midrule
        & $\omega_{1}$ & 0.097 & action smoothing\\
        & $\omega_{2}$ & 1.579 & number of active muscles above 15\% activity\\
        & $\omega_{3}$ & 0.131 & joint limit torque \\
        & $\omega_{4}$ & 0.073 & GRFs above 1.2 BW\\
        \midrule
        & $\Delta\alpha$ & $9\times 10^{-4}$ & change in adaptation rate \\
        & $\theta$ & 1000 & performance threshold \\
        & $\beta$ & 0.8 & running avg. smoothing\\
        & $\lambda$ & 0.9 & decay term\\
        \bottomrule
        \end{tabular}
    \end{subtable}\\
    \vspace{0.5cm}
    \end{table*}

\end{document}